# Digital Rock Typing DRT Algorithm Formulation with Optimal Supervised Semantic Segmentation

Omar Alfarisi, Djamel Ouzzane, Mohamed Sassi, and TieJun Zhang

**Abstract**

Each grid block in a 3D geological model requires a rock type that represents all physical and chemical properties of that block. The properties that classify rock types are lithology, permeability, and capillary pressure. Scientists and engineers determined these properties using conventional laboratory measurements, which embedded destructive methods to the sample or altered some of its properties (i.e., wettability, permeability, and porosity) because the measurements process includes sample crushing, fluid flow, or fluid saturation. Lately, Digital Rock Physics (DRT) has emerged to quantify these properties from micro-Computerized Tomography (uCT) and Magnetic Resonance Imaging (MRI) images. However, the literature did not attempt rock typing in a wholly digital context. We propose performing Digital Rock Typing (DRT) by: (1) integrating the latest DRP advances in a novel process that honors digital rock properties determination, while; (2) digitalizing the latest rock typing approaches in carbonate, and (3) introducing a novel carbonate rock typing process that utilizes computer vision capabilities to provide more insight about the heterogeneous carbonate rock texture.

## Introduction

If rock typing (*1-4*) determines the critical aggregation of a geological formation's physical and chemical properties (*5-18*), then the path to efficient determination would be a digital one assisted by artificial intelligence. The latest rock typing approaches, Conjunction Rock Properties Convergence (CROPC) (*1*) and Rock Matrix (RocMate), (*19*) have proven to integrate three conventional methods, the Rock Lithology Facies (Lithofacies), Rock Electrically Measured Properties Facies (Electrofacies), and Rock Petrophysical Properties Facies (Petrofacies) (*20*). Although scientists and engineers have performed lots of work using these five approaches, it was without digital warping. Here, we show a novel integration aided by the digital world capabilities, mainly Machine Learning to produce a new process, Digital Rock Typing DRT. After the digital system's demonstration of quantifying: (1) Lithology (*21*), (2) Pore Throat Network (*18*), (3) Permeability (*18*), and (4) Capillary Pressure (*18*), we realized it is the right time to launch, for the first time, the DRT.

Geoscientists and engineers need appropriate rock types to build the 3D geological and flow simulation models. Any work involving dynamic properties determination remains incomplete without proper rock types as they hold the physical and chemical properties boundaries for future flow behavior predictions. Therefore, with DRT, the scientific and industrial communities can generate rock types faster, higher quality for a large mass of data and consistently.

**The Beginning of Rock Typing**

In the '50s of the 20th century (*2*), Archie developed his famous rock typing diagram that remained the primary rock typing reference for almost half a century. Figure 1 displays the exact original version of what Archie built.

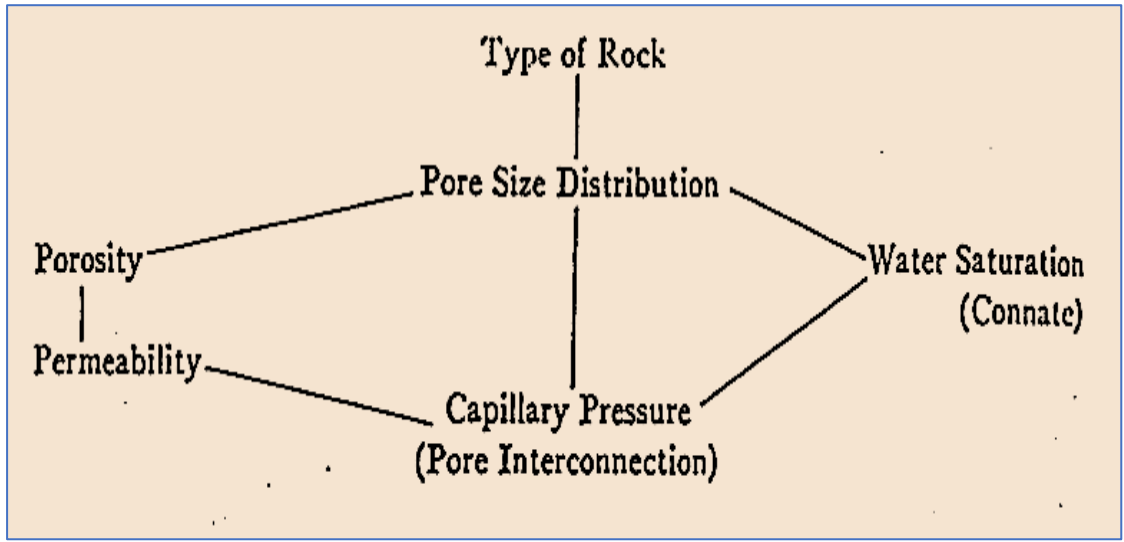

*Figure 1. Type of Rock (reproduction from Archie -1950) [1].*

After exactly 59 years from this first Rock Typing method, a new Rock Typing process, CROPC, saw the light (*1*). The following section discusses removing porosity (of Archie's rock typing) because porosity alone does not represent carbonate rock dynamic properties. Carbonate rock heterogeneous morphology contains an intraparticle pore system related to carbonate bioclast diagenesis. However, carbonate porosity has a role in permeability determination if the morphology of the rock is known. Therefore, decoding morphology (*18*) is one of Archie's rock typing missing elements. In other words, in carbonate, the porosity is just secondary support of the main features that control rock typing: Lithology, Permeability, and Capillary Pressure. Therefore, in the subsequent sections, we describe the efforts after Archie to develop the Carbonate Rock Typing.

# The Carbonate Rock Typing After Archie

## 1) The Dunham Carbonate Rock Typing - 1961

After 11 years of Archie's Rock Typing (*3*), Dunham introduced rock typing based on Depositional Texture. This work identifies the ranges of grain size percentage in the rock type to define a rock type. The classes are Mudstone, Wackestone, Packstone, Grainstone, and Bondstone. We display these classes in Figure 2, which shows a reproduction of the original work as produced by Dunham (*3*).

*Figure 2. Deposition Based Carbonate Rock Classification (reproduction from (3)).*

Dunham's classification has neither used nor described diagenesis because it depends on the depositional environment. However, the word "diagenesis" appears in Figure 2 under the "Depositional Texture Not Recognizable" section. From which we inferred the need for further research in Diagenesis.

## 2) The Lucia Carbonate Rock Typing – 1995

After 45 years of Archie's rock typing (*2*), Lucia discussed his previous work in 1983 that considered diagenesis. The main driver for Lucia's rock typing was the interparticle porosity and its relationship to Permeability, as per Figure 3, which is we reproduce from the original paper (*4*). This interparticle porosity has one major drawback, which is also evident in Figure 3 is that porosity and permeability in carbonate need the pore throat size network (22) to link them. Lucia (4) has mentioned in the paper that he used a similar concept to Archie's work (2), with the inherited challenges.

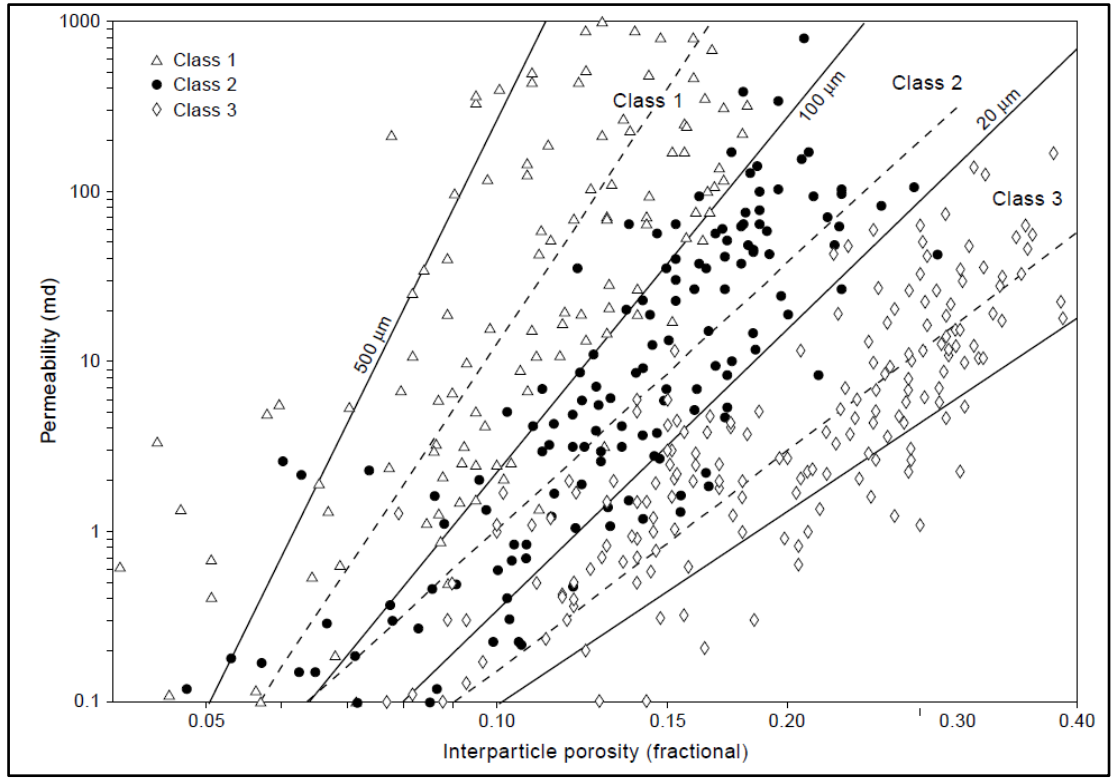

*Figure 3. Rock Typing based on Interparticle Porosity (reproduction from (4)).*

We notice that Lucia introduced the effect of Diagenesis, where he described vug types; however, the link was to interparticle porosity only. Lucia's rock typing produces rock typing that does not honor rock dynamic properties, leading to limited usage of Lucia's rock types for flow simulation purposes. Besides, Lucia (4) has introduced the link between $P_c$ and rock types of the unimodal and bimodal pore system. However, the distinction between the two $P_c$ curves that honor unimodal and bimodal has no description for the differences.

We understand why Lucia could not show clear differentiation because Lucia's plot displayed two changes at the same time; these two changes are:

1- Pore throat size for the two different rocks has two distinctive peaks.
2- One core sample is unimodal, and the other is bimodal.

Therefore, Lucia's work has two interlinked difficulties. One is due to using porosity instead of the pore throat size. The second is that the $P_c$ curve of Lucia represents more than one rock type. Having these two unsolved makes it hard to perform rock typing in carbonate beyond 1995. However, this opened the door for further research for a solution in 2009 by CROPC (*1*) to solve one of the unsolved points (pore throat size rather than porosity). At the same time, RocMate (*19*) has solved the second unsolved difficulty by making one change at a time, rather than two changes, which we explain further in subsequent sections.

## 3) The CROPC Carbonate Rock Typing - 2009

After 59 years of Archie's Rock Typing (*1*), the research continued to solve the difficulty in Lucia. CROPC introduced a new concept in rock typing to solve the considerable overlap between classes and the overlap in rock physical properties, as shown in Figure 4, which is a reproduction from the original paper (*1*).

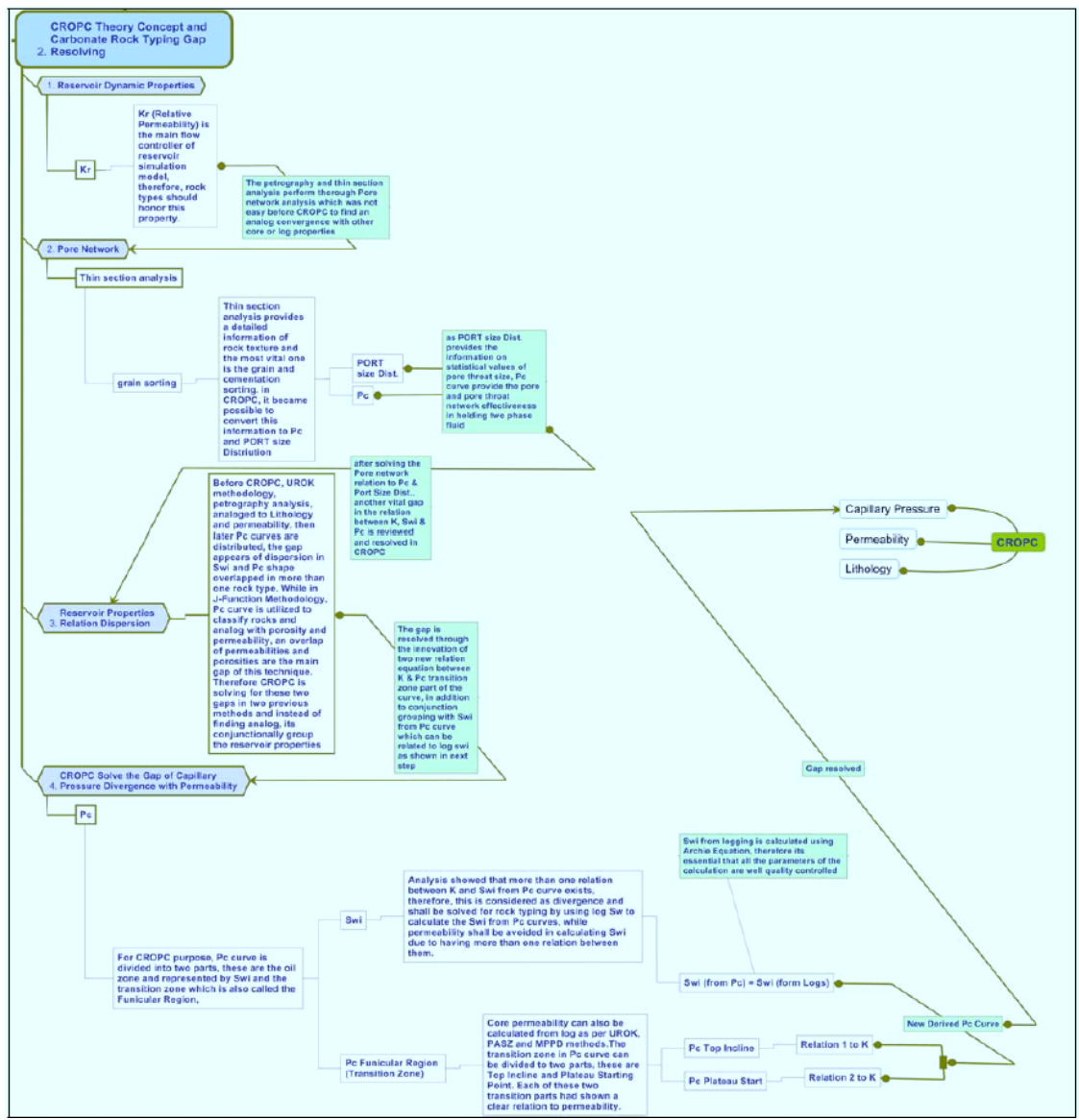

*Figure 4. CROPC Carbonate Rock Typing based on Rock Properties Conjunction (reproduction from (1)).*

In CROPC, the rock type definition is in four words, so rock type is rocks with similar properties. This Conjunction of properties prevented the significant overlap of Lucia. Therefore, CROPC produced segregated properties with minimal overlap. Figure 5 shows the CROPC method flow diagram reproduced from the original paper (*19*).

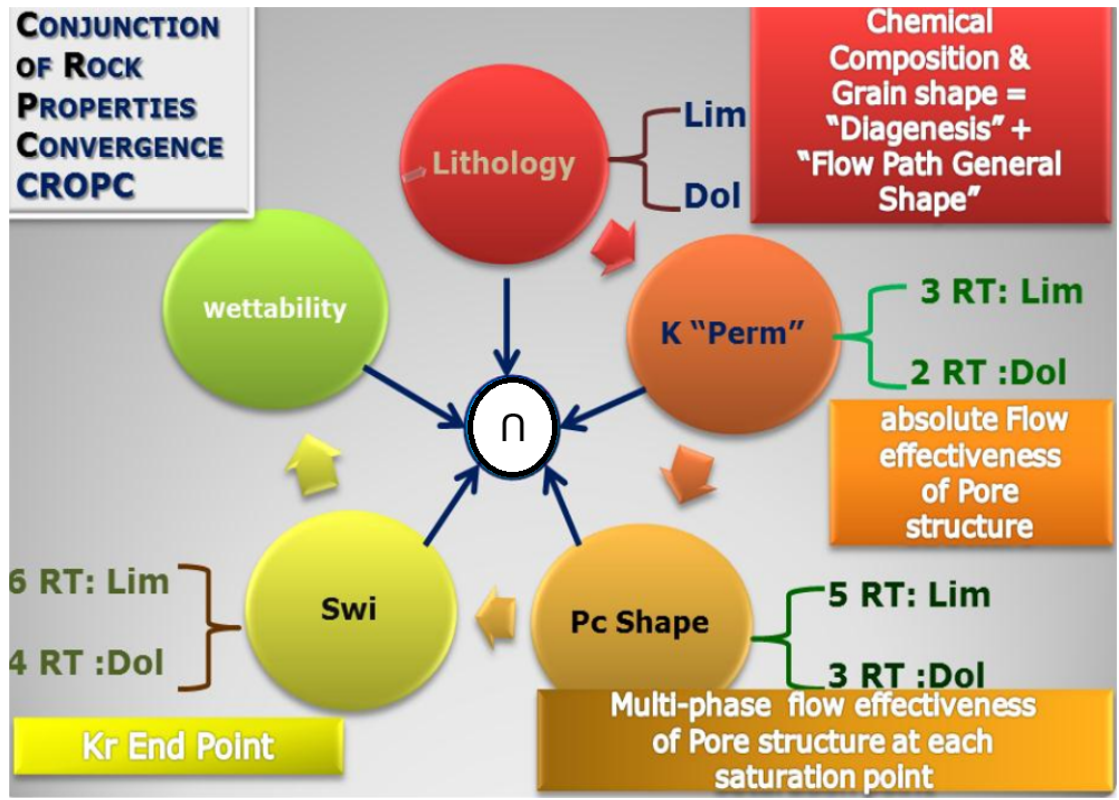

*Figure 5. Rock Typing Conjunction of Properties "CROPC" (reproduction from (19)).*

For dynamic flow simulation, CROPC offered a great advantage as a static properties classifier, especially in producing rock types that have no permeability overlap among different rock types. However, linking rock types to unimodal and bimodal was not part of the CROPC scope, making it hard for geoscientists to interpolate data between wells in the 3D geological model. Therefore, the need for further research to resolve this difficulty became a necessity that we explain in the next section on how another new rock typing, RocMate, solved this difficulty.

### 4) RocMate Rock Typing - 2013

After 63 years of Archie's rock typing (*14, 19*), RocMate worked to resolve for including the unimodal and bimodal in the Classification. In Table 1, the pore models were the first to be classified, producing the first round (stage) of classification based on Singular (unimodal), Dual (bimodal), and Triple (trimodal) pore systems. Then the second round established the link between permeability and $P_c$ using two points in the $P_c$ curve, as described in morphology decoder (22). While Table 2, which is a reproduction from the original paper (19), shows the results of the rock typing catalog, including the range of $P_c$ (in psi units), Permeability(k) (in mD (milli-Darcey) units), and $S_{wi}$ (Intrinsic Water Saturation (fraction)).

Table 1. Rock Typing based on Uni-Bi-Tri "RocMate" and "CROPC" (reproduction from (19)).

|  | Micro | Meso | Macro | % | % | % |
|---|---|---|---|---|---|---|
| Triple | O | O | O | 20-50-30 | 20-30-50 |  |
| Dual | O |  | O | 25-75 | 50-50 | 75-25 |
| Dual | O | O |  | 25-75 | 50-50 | 75-25 |
| Dual |  | O | O | 25-75 | 50-50 | 75-25 |
| Singular | O |  |  | 100 |  |  |
| Singular |  | O |  |  | 100 |  |
| Singular |  |  | O |  |  | 100 |

Table 2. Carbonate Rock Typing Catalogue of "RocMate" (reproduction from (19)).

| Lim Rock | New RT L # # # Lim Perm Pc Swi | Perm Range | PC shape | Swi |
|---|---|---|---|---|
| 1 | L111 | >60 | note2&3 (<400,<100psi) | note 4 &5 (7-13%) |
| 2 | L121 | >60 | note2&3 (>400, <100psi) | note 4 &5 (7-13%) |
| 3 | L231 | 7<K<60 | Note 6 & 7 (<700, >80psi) | Note 8&9 (7-13%) |
| 4 | L241 | 7<K<60 | Note 6 & 7 (>700,<30psi) | Note 8&9 (7-13%) |
| 5 | L242 | 7<K<60 | Note 6 & 7(>700,<30psi) | Note 8&9 (14-20%) |
| 6 | L351 | 1<K<7 | Note 10&11 (<1100,<100psi) | Note 12&13 (7-13%) |
| 7 | L352 | 1<K<7 | Note 10&11 (<1100,<100psi) | Note 12&13 (14-20%) |
| 8 | L361 | 1<K<7 | Note 10&11 (<1400, >250psi) | Note 12&13 (7-13%) |
| 9 | L372 | 1<K<7 | Note 10&11 (>1400, >100psi) | Note 12&13 (14-20%) |
| 10 | L373 | 1<K<7 | Note 10&11 (>1400, >100psi) | Note 12&13 (21-27%) |
| 11 | L374 | 1<K<7 | Note 10&11 (>1400, >100psi) | Note 12&13 (28-34%) |
| 12 | L382 | 1<K<7 | Note 10&11 (>1400, >100psi) | Note 12&13 (14-20%) |
| 13 | L461 | 0.1<K<1 | {P(<1000, <400)} | P (7-13%) |
| 14 | L492 | 0.1<K<1 | {Q(<1200,<600)} | Q (14-20%) |
| 20 | LD5 | <0.1 |  |  |

While in Figure 6, we see different pore systems (Single, Dual, Triple) link the permeability to porosity, which is a reproduction from (*1, 19*).

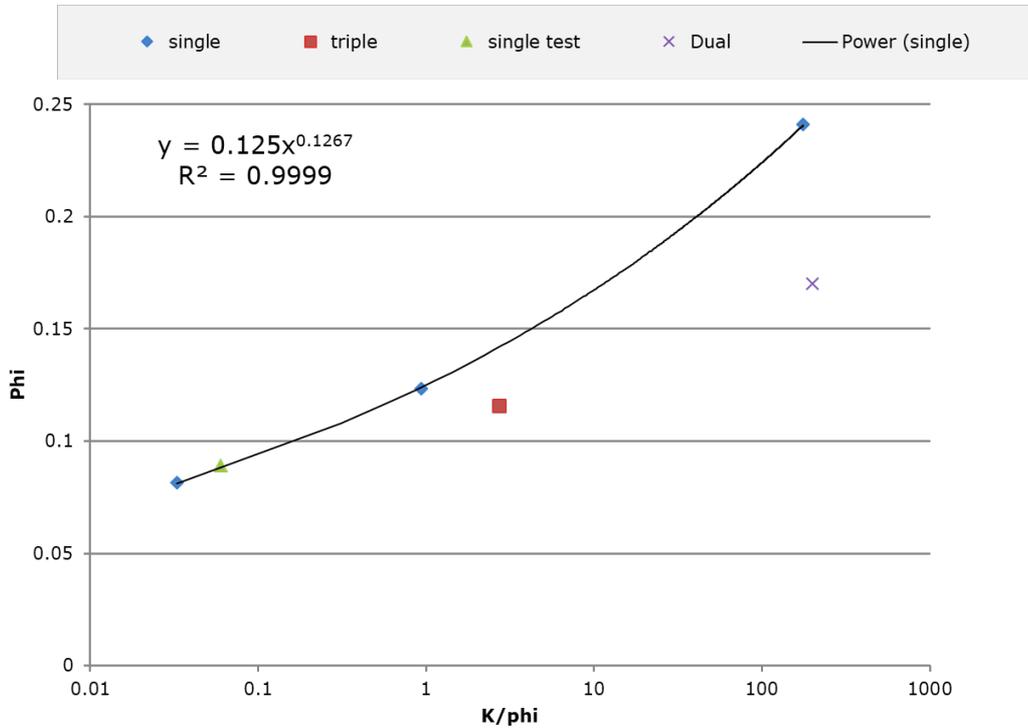

*Figure 6. Porosity and Pore Systems Link to permeability (reproduction from (19)).*

With all these advances in solving 20th-century rock typing challenges (*1, 19*), what we notice still missing is machine learning utilization. Machine Learning algorithms recognized the rock pore systems using Morphology Decoder (*22*) and semantically segmented the different heterogeneous regions in the carbonate rock to produce the Pore Throat Size Network (PorThN). This advancement led us to develop a modern age framework for Digital Rock Typing (DRT).

## Digital Rock Typing (DRT) - 2019

After 69 years of Archie's rock typing, we introduce Digital Rock Typing (DRT) in this paper. We explain the DRT algorithm in Figure 7. With the digital aid, we provided a framework to scale up the identification of each rock segment shown in Figure 8. CMV to produce its properties and to deliver the rock type classification. We introduced a new digital label of rock morphology features (i.e., The Rhombohedral Configuration, Intragranular Vugs Connectivity, and Cementation) using deep learning, which is vital to feed these parameters as input to Rock Classification. DRT produces, for the first time, rock types that depend on pore throat networks using machine learning. We created the digital rock type identifier chart, shown in Figure 9, and we called it the Carbonate Morphology (CAMO) chart. This chart ensured porosity-permeability-morphology relation classifying the connected, non-connected, and micropore morphology.

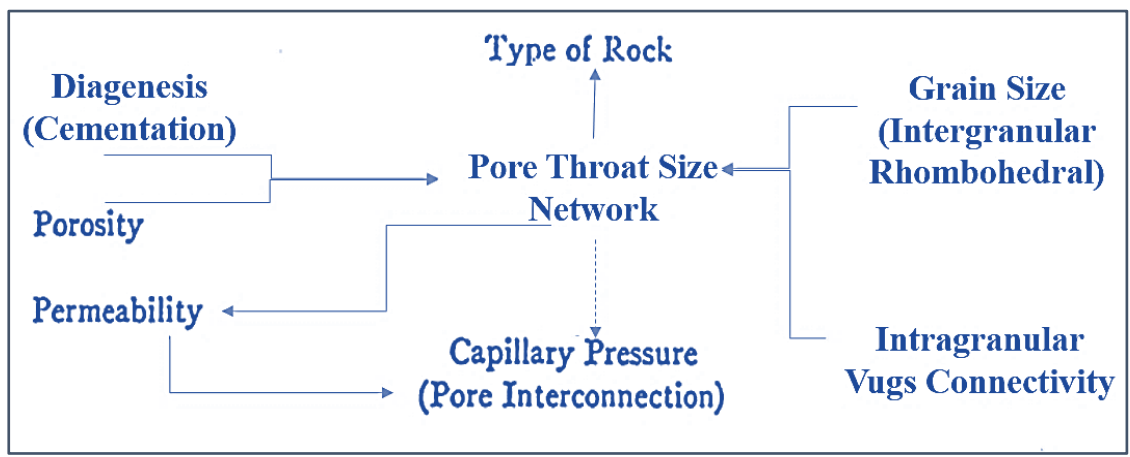

*Figure 7. Digital Rock Typing DRT Algorithm Diagram.*

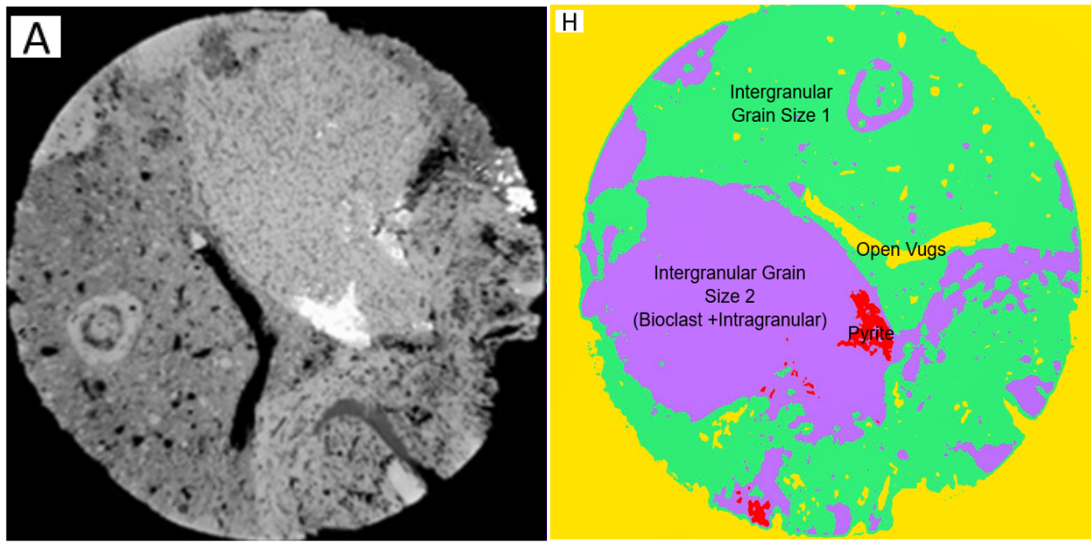

*Fig. 7. Morphology Decoder, a Rock Texture Classifier with Supervised Semantic Segmentation of Difference-of-Gaussian Random Forest Algorithm. (A) The original 3D µCT image with 28 um resolution. (H) Morphologies with their corresponding morphology, recognized by the machine as different zones in Cretaceous carbonate. Each morphology has a different impact on the flow property of a natural fluid. This figure is a reproduction of the original paper (22).*

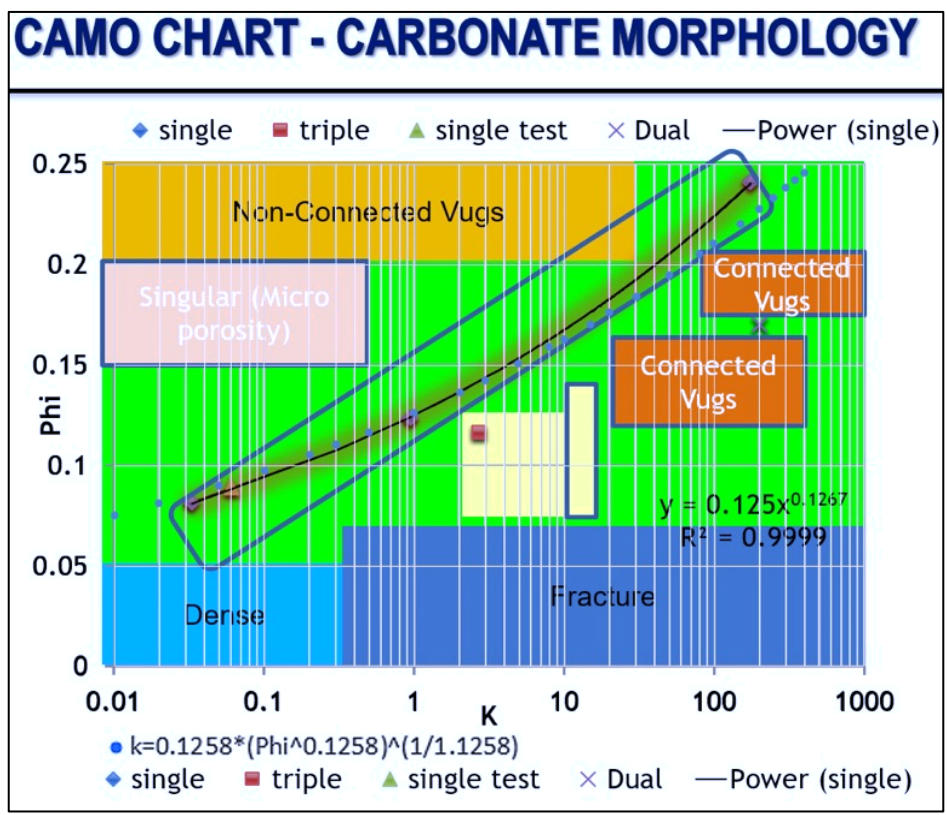

*Figure 9. CAMO Chart upgrades k(mD) vs. Porosity(fraction) of (19) as our research outcome.*

**DRT Algorithm Steps Summary**

The DRT demonstrates that the machine can determine the rock types in the logging and core domains. We describe each of the components of the DRT Algorithm below:

1. Porosity and lithology:

    In the logging domain, porosity logging tools determine porosity and lithology. While, in the core domain, the machine performs one of the two methods:

    a. Image analysis with Image Resolution Optimized Gaussian Algorithm (IROGA) (*17, 23*).

    b. Machine Learning Difference of Gaussian  Random Forest Algorithm MLDGRF (*17, 23*).

2. Permeability: The permeability determination is achievable by the machine in two steps in the logging (MRI) and core domain ($\mu$CT and MRI):

    a. Calibrate MRI and $\mu$CT PorThN with image intensity (*18*).

    b. Machine learning, MLDGRF algorithm to classify Singular, Dual, and Triple pore systems (*18*).

3. Capillary Pressure Curve: The $P_c$ determination is achievable by the machine for the logging and core domain in two steps:

    a. The machine calculates $P_{cd}$ from permeability using the Capillary-Permeability-Function (P-Function) (*24*).

    b. The machine uses $P_{cd}$ and $P_{cu}$ points in the P-function power function to calculate the $P_c$ curve (*24*).

4- <u>Digital Rock Typing:</u> The machine uses the findings of this research, Carbonate Morphology (CAMO) chart, shown in Figure 9, to determine the rock type.

**Conclusions**

1- The DRT algorithm works for both logging and core domains because all the required features for DRT are determinable using machine learning and automated processes available for both environments.

2- DRT's contribution, on top of the existing wisdom, is that the machine can determine rock types with the novel CAMO chart that established the interlink between permeability-porosity-PorThN.

**Future Work**

We recommend researching and testing the rock electrical properties, relative permeability, wettability range, and alteration with artificial intelligence.

**Affiliation**
Omar Alfarisi (ADNOC Offshore), Djamel Ouzzane (ADNOC), Mohamed Sassi (Khalifa University), and TieJun Zhang (Khalifa University).

**Acknowledgment**
The authors thank the support and encouragement received from ADNOC, ADNOC Offshore, and Khalifa University of Science and technology. We express our appreciation to Mr. Yasser Al-Mazrouei, Mr. Ahmed Al-Suwaidi, Mr. Ahmed Al-Hendi, Mr. Ahmed Al-Riyami, Mr. Andreas Scheed, Mr. Hamdan Al-Hammadi, Mr. Khalil Ibrahim, Mr. Mohamed Abdelsalam, Dr. Ashraf Al-Khatib, Prof. Isam Janajreh, Dr. Aikifa Raza, Dr. Hongxia Li, and Mr. Hongtao Zhang.

References
1. O. Al-Farisi, M. Elhami, A. Al-Felasi, F. Yammahi, S. Ghedan, in *SPE/EAGE Reservoir Characterization & Simulation Conference*. (European Association of Geoscientists & Engineers, 2009), pp. cp-170-00073.
2. G. E. Archie, Introduction to petrophysics of reservoir rocks. *AAPG bulletin* **34**, 943-961 (1950).
3. R. J. Dunham, Classification of carbonate rocks according to depositional textures.  (1962).
4. F. J. Lucia, Rock-fabric/petrophysical classification of carbonate pore space for reservoir characterization. *AAPG bulletin* **79**, 1275-1300 (1995).
5. E. BinAbadat *et al.*, in *SPE Reservoir Characterisation and Simulation Conference and Exhibition*. (OnePetro, 2019).
6. O. Al-Farisi, A. Belgaied, H. Shebl, G. Al-Jefri, A. Barkawi, Well Logs: The Link Between Geology and Reservoir Performance. *Abstract Geo2002* **96**,  (2002).
7. O. Al-Farisi *et al.*, in *Abu Dhabi International Conference and Exhibition*. (OnePetro, 2004).
8. T. Weldu, S. Ghedan, O. Al-Farisi, in *SPE Production and Operations Conference and Exhibition*. (OnePetro, 2010).
9. O. Al-Farisi, S. Budebes, T. Hamdy, O. Y. Al-Sheehe, in *SPE Reservoir Characterisation and Simulation Conference and Exhibition*. (Society of Petroleum Engineers, 2011).
10. S. Ghedan, T. Weldu, O. Al-Farisi, in *Abu Dhabi International Petroleum Exhibition and Conference*. (OnePetro, 2010).
11. S. Khemissa *et al.*, in *SPE Reservoir Characterisation and Simulation Conference and Exhibition*. (OnePetro, 2011).
12. S. Budebes *et al.*, in *SPE Reservoir Characterisation and Simulation Conference and Exhibition*. (OnePetro, 2011).
13. O. Al-Farisi *et al.*, in *SPE Reservoir Characterisation and Simulation Conference and Exhibition*. (OnePetro, 2011).
14. O. Al-Farisi, *Carbonate rock matrix, RocMate: A novel static modeling technique*.  (The Petroleum Institute (United Arab Emirates), 2012).
15. H. Li, O. AlFarisi, B. Voort, C. Dimas, T. Zhang, in *Abu Dhabi International Petroleum Exhibition & Conference*. (OnePetro, 2016).
16. C. Carpenter, Machine-Learning Image Recognition Enhances Rock Classification. *Journal of Petroleum Technology* **72**, 63-64 (2020).
17. O. Al-Farisi *et al.*, in *SPE Reservoir Characterisation and Simulation Conference and Exhibition*. (OnePetro, 2019).
18. O. Alfarisi *et al.*, Morphology Decoder: A Machine Learning Guided 3D Vision Quantifying Heterogenous Rock Permeability for Planetary Surveillance and Robotic Functions. *arXiv preprint arXiv:2111.13460*,  (2021).
19. O. Al-Farisi *et al.*, in *International Petroleum Technology Conference*. (OnePetro, 2013).
20. J. A. Rushing, K. E. Newsham, T. A. Blasingame, in *SPE Unconventional Reservoirs Conference*. (OnePetro, 2008).
21. O. Alfarisi *et al.*, Machine Learning Guided 3D Image Recognition for Carbonate Pore and Mineral Volumes Determination. *arXiv preprint arXiv:2111.04612*,  (2021).
22. O. Alfarisi *et al.* (TechRxiv, 2021).
23. O. Alfarisi, Z. Aung, M. Sassi, Deducing of Optimal Machine Learning Algorithms for Heterogeneity. *arXiv preprint arXiv:2111.05558*,  (2021).
24. O. Alfarisi, D. Ouzzane, M. Sassi, T. Zhang, The Understanding of Intertwined Physics: Discovering Capillary Pressure and Permeability Co-Determination. *arXiv preprint arXiv:2112.12784*,  (2021).